\documentclass[letterpaper, 10 pt, conference]{ieeeconf}  % Comment this line out if you need a4paper

\IEEEoverridecommandlockouts                              % This command is only needed if 
\usepackage{times}
\usepackage{epsfig}
\usepackage{graphicx}
\usepackage{amsmath}
\usepackage{amssymb}
\usepackage{comment}
\usepackage{footnote}
\usepackage{epstopdf}
\usepackage{comment}
\usepackage{multirow}
\usepackage{algorithm}
\usepackage{algorithmic}
\usepackage{times}
\usepackage{epsfig}
\usepackage{graphicx}
\usepackage{amsmath}
\usepackage{amssymb}
\usepackage{wasysym}
\usepackage[ruled,vlined,algo2e]{algorithm2e}
\usepackage{color}
\usepackage{dsfont}	

\usepackage{adjustbox}

\usepackage{dblfloatfix}    % To enable figures at the bottom of page

\usepackage{authblk}
\usepackage{romannum}
\usepackage[table,xcdraw]{xcolor}
\usepackage{gensymb}
\usepackage{graphicx} % [demo] option for empty figure
\usepackage{hyperref}

\title{\LARGE \bf Accelerating Point Cloud Ground Segmentation: From Mechanical to Solid-State Lidars}

\author{Xiao Zhang, Zhanhong Huang, Antony Garcia, and Xinming Huang% <-this % stops a space
\thanks{This work was supported in part by The MathWorks Inc.}% <-this % stops a space
\thanks{X. Zhang, Z. Huang, A. Garcia, and X. Huang  are with Worcester Polytechnic Institute, 100 Institute Road, Worcester, MA 01609, USA
    {\tt\small \{xzhang25, zhuang5, agarcia3, xhuang\}@wpi.edu}}%
    
}

\begin{document}

\maketitle
\thispagestyle{empty} 
\pagestyle{empty}

%%%%%%%%%%%%%%%%%%%%%%%%%%%%%%%%%%%%%%%%%%%%%%%%%%%%%%%%%%%%%%%%%%%%%%%%%%%%%%%%
\begin{abstract}
In this study, we propose a novel parallel processing method for point cloud ground segmentation, aimed at the technology evolution from mechanical to solid-state Lidar (SSL). We first benchmark point-based, grid-based, and range image-based ground segmentation algorithms using the SemanticKITTI dataset. Our results indicate that the range image-based method offers superior performance and robustness, particularly in resilience to frame slicing. Implementing the proposed algorithm on an FPGA demonstrates significant improvements in processing speed and scalability of resource usage. Additionally, we develop a custom dataset using camera-SSL equipment on our test vehicle to validate the effectiveness of the parallel processing approach for SSL frames in real world, achieving processing rates up to 30.9 times faster than CPU implementations. These findings underscore the potential of parallel processing strategies to enhance Lidar technologies for advanced perception tasks in autonomous vehicles and robotics. The data and code will be available post-publication on our GitHub repository: \url{https://github.com/WPI-APA-Lab/GroundSeg-Solid-State-Lidar-Parallel-Processing}

% The contributions of this work not only set a new standard for Lidar ground plane segmentation but also underscore the potential of FPGA-based systems in the field of autonomous driving technologies. By offering a comprehensive solution that combines methodological innovation with practical application and deployment, this study paves the way for future advancements in autonomous vehicle perception systems.

\end{abstract}

% \footnote{This work was supported in part by The MathWorks Inc. X. Zhang is with Department of Electrical and Computer Engineering, Worcester Polytechnic Institute, Massachusetts 01609, USA. (e-mail:xzhang25@wpi.edu).

% W. Jachimczyk is with the Computer Vision and Autonomous Vehicles Team, The MathWorks, Massachusetts, 01760, USA (e-mail:wjachimc@mathworks.com).

% Z. Huang, A. Garcia, and X. Huang are with Department of Electrical and Computer Engineering, Worcester Polytechnic Institute, Massachusetts 01609, USA.  (e-mail:\{zhuang5, agarcia3, xhuang\}@wpi.edu). 
% }

%%%%%%%%%%%%%%%%%%%%%%%%%%%%%%%%%%%%%%%%%%%%%%%%%%%%%%%%%%%%%%%%%%%%%%%%%%%%%%%%

\section{Introduction}

Over the past decade, Lidar technology has evolved significantly, marked by enhancements in sensors that utilize 905 nm (near-infrared) and 1550 nm (shortwave infrared) wavelengths. The 905 nm wavelength is favored for its cost-effectiveness and energy efficiency, while the 1550 nm wavelength offers enhanced safety for human eyes and better performance under adverse conditions.

Lidar's precision in three-dimensional depth sensing, which surpasses that of traditional radar with centimeter to millimeter wavelengths, has facilitated its broad adoption in fields such as infrastructure monitoring \cite{Zhang_Xiao_Coifman_Mills_2020}, robotics \cite{bogoslavskyiFastRangeImagebased2016}, and autonomous vehicles \cite{zermas2017fast} \cite{zhou2021panoptic}. The traditional mechanical Lidar sensors, which require rotation to achieve a 360-degree field of view, often pose challenges in terms of size, cost, and reliability. These challenges have spurred the development of solid-state Lidar (SSL) systems, which are categorized into flash-based and scanning-based (semi-SSL) technologies, with the latter often employing Micro-Electro-Mechanical Systems (MEMS) mirrors to reduce mechanical components and enhance resolution \cite{liProgressReviewSolidState2022a}.

\newpage
Despite advancements, the transition from mechanical to solid-state Lidar introduces significant challenges, particularly in adapting Lidar algorithms for these new systems—a topic that remains under-researched. Moreover, the increased resolution of modern Lidar systems poses challenges for real-time processing on embedded platforms. \cite{zhangRealTimeFastChannel2022}\cite{lyuChipNetRealTimeLidar2019}.

This paper explores how the intrinsic differences between mechanical and solid-state Lidar sensors impact existing Lidar perception algorithms. Focusing on ground segmentation—a crucial classification task—we investigate and validate the potential for parallelism to enhance performance and hardware efficiency of point cloud processing across various Lidar sensors. Our major contributions are summarized as follows:

% \textbf{Exploration of Frame Segmentation-Based Parallel Processing on Mechanical Lidar Sensors}: We developed and validated parallel processing techniques for point cloud processing using point-based, voxel-based, and range-image-based methods with the SemanticKITTI dataset. Our results highlight the robustness of the range-image method, particularly against slicing. Despite increases in slice numbers, the range-image-based method maintained consistent performance, outperforming others with stable IoU and F1 scores.

% \textbf{Adoption and Verification of Parallel Computation for solid-state Lidar Sensors}: We adapted the range-image-based ground segmentation method to SSL using a custom dataset from a synchronized Camera-SSL framework. Our experiments validated a frame-slice-based parallel ground segmentation framework, demonstrating its precision and robustness, with performance customizable through threshold adjustments.

% \textbf{Innovative Implementation of Range-Image Ground Segmentation on FPGA for SSL}: We were the first to implement range-image ground segmentation on FPGA for SSL sensor, utilizing both parallel and sequential strategies. Our comparative analyses revealed substantial resource efficiencies through frame-slice paralleling. In setups with multiple processing units, we achieved speed-ups of up to 30.9 times compared to CPU implementations by additional resource allocation.

\textbf{Comparison of existing ground segmentation algorithms under frame-slice using the SemanticKITTI dataset.}
We investigated frame-wise parallelization techniques for point cloud processing using point-based, voxel-based, and range-image-based 
 ground segmentation methods with the SemanticKITTI dataset.  Our experiments demonstrated the robustness of the range-image-based method, which maintained consistent performance despite increased slicing, outperforming the other methods in overall performance and stability.

\textbf{Design of a scalable ground segmentation acceleration architecture for solid-state Lidar.}
To validate the hardware efficiency of our proposed method, we developed a scalable, parallel ground segmentation accelerator tailored for solid-state Lidar and implemented it on an FPGA platform. The architecture was designed to support a flexible and modular configuration, enabling it to adapt to different resource constraints and application requirements.  Comprehensive analyses demonstrated resource efficiency, high throughput, and low power consumption.

\textbf{Validation of our proposed approach and architecture using our custom dataset from a test Vehicle.}
Building on our previous designs, we evaluated the proposed hardware-efficient pipeline using a custom dataset obtained from a calibrated and synchronized camera-SSL framework. Both quantitative and qualitative evaluations confirmed the robustness and effectiveness of our frame-wise parallel ground segmentation framework for SSL. In multi-processing unit configurations, the accelerator maintained accuracy and achieved up to a 30.9× speedup compared to CPU implementations.

% \begin{figure}[h]
%     \includegraphics[width=\linewidth]{figures/ISICAS24_intro_demo2.pdf}
%     \caption{Ground Segmentation Results of T-Junction and Crossway Scenarios from the SemanticKITTI Dataset}
%     \vspace{-0.4cm}
%     \label{fig:ISICAS24_intro_demo}
% \end{figure}

% \begin{figure*} [h]
%     \begin{center}
%         \includegraphics[width=\linewidth]
%         % \includegraphics[width=\linewidth, height=4.5 cm]
%         {figures/ISICAS24_alg_pipeline.pdf}
%         \caption{Pipeline of Fast Channel-based Ground segmentation: This diagram illustrates the process from extracting range images from organized point clouds to segmentation, with frame sizes determined by the Lidar sensor's vertical and horizontal resolutions (e.g., 64x2048 for OS-64 Lidar)}
%         \label{fig:ISICAS24_alg_pipeline}
%     \end{center}
% \end{figure*}

% (1) We propose a novel, range image based, three-pass grouping algorithm called fast channel clustering. To address the problem of over-segmentation caused by missing points, we design a scalable connection filter to 
%  search the local neighborhood for inter-channel connections. A merge table is used to store the connected labels efficiently.

% (2) We implement and verify the FCC algorithm on a Zynq-7000 FPGA. Combining a line buffer with a unique sort network that enables fully parallel calculations of Euclidean distances and unique sorting among the neighboring points, the FPGA implementation achieves 2 to 3 orders of magnitude of speed-up over CPU run-time.

\section{Related Work}
\subsection{Evolution towards Solid-State Lidar} 
Although solid-state Lidar (SSL) technology has been in development for decades, it is only in recent years that products with adequate field of view (FoV) have become commercially viable. Among these, the Livox series MEMS SSL stands out for its cost-efficiency. The Mid-40s model, for instance, features a detection range of 260 meters, performing on par with a 32-line mechanical Lidar operating at 10 Hz. Significant advancements in SSL technology include Cui et al.'s development of an automatic calibration system for industrial-grade SSLs \cite{cuiACSCAutomaticCalibration2020a} and Peng et al.'s creation of a 3D object detection and tracking system designed for automotive-grade SSLs \cite{peng3DObjectsDetection2023}. Much of the existing research on SSL has focused on Simultaneous Localization and Mapping (SLAM) applications, with notable contributions from Wei et al. \cite{weiEnhancingSolidState2021} and Li et al. \cite{liHighPerformanceSolidStateLidarInertialOdometry2021}, who proposed enhanced mapping systems and lightweight odometry solutions optimized for urban SLAM and feature extraction tasks.

Despite the affordability and innovative non-repetitive scanning patterns of SSLs, challenges remain in adapting these systems due to their unique circular projection shape. On the other hand, regular frame MEMS SSLs, which offer more conventional rectangular scanning frames, have seen increasing adoption in autonomous driving applications. However, detailed studies exploring the full utility and performance of these systems are still limited in the literature.

\subsection{Lidar Ground Segmentation}

While learning-based methods are prevalent in current evaluations, their effectiveness is hampered by the scarcity of data and the extensive labeling labor required for adapting the methods for SSL sensor.  Consequently, we surveyed traditional Lidar point cloud ground segmentation methods.  Ground segmentation algorithms for Lidar data processing are traditionally categorized into point-based \cite{lim2022patchworkconcentriczonebasedregionwise} \cite{fischlerRandomSampleConsensus1981} \cite{chenGaussianProcessBasedRealTimeGround2014}, grid-based \cite{douillardHybridElevationMaps2010} \cite{pingelImprovedSimpleMorphological2013}, and range-image-based methods \cite{bogoslavskyi2017efficient}. 

The point-based approaches only rely on unordered single points, utilizing statistical \cite{fischlerRandomSampleConsensus1981} or other models \cite{lim2022patchworkconcentriczonebasedregionwise} for data fitting, with Random Sample Consensus (RANSAC) \cite{fischlerRandomSampleConsensus1981} and Gaussian-based \cite{chenGaussianProcessBasedRealTimeGround2014} methods as typical examples. Grid-based methods, on the other hand, resample sparse point clouds into grids. \cite{douillardHybridElevationMaps2010} adapted the mean height within a Bird's Eye View (BEV) grid segment as a proxy for point cloud representation, employing gradient-based classification to isolate ground points. Pingel et al. \cite{pingelImprovedSimpleMorphological2013} introduced an improved Simple Morphological Filter (SMRF) for ground point segmentation using a linearly increasing window and simple slope thresholds. Range-image-based methods, such as those detailed by Bogoslavskyi et al. \cite{bogoslavskyi2017efficient}, convert point clouds into 2D images where segmentation algorithms can be more straightforwardly applied, often resulting in more efficient processing and potentially more accurate ground detection.

\begin{table*}[t]
\begin{center}
\label{tb:evaluation_semantic}
\caption{Performance Evaluation Ground Segmentation Via Sub-frame Number on semantickKITTI}
 \begin{adjustbox}{width=14cm, center}
\begin{tabular}{c|cc|cc|cc}
\hline
\multicolumn{1}{l|}{\textbf{Method (Type)}} & \multicolumn{2}{c|}{\textbf{SMRF (Grid-based)}} & \multicolumn{2}{c|}{\textbf{RANSAC (Point-based)}} & \multicolumn{2}{c}{\textbf{Depth Ground (Range image-based)}} \\ \hline
\textbf{Subframe Num}                   & \textbf{IoU}     & \textbf{F1}     & \textbf{IoU}      & \textbf{F1}      & \textbf{IoU}        & \textbf{F1}         \\ \hline \hline
1                                    & 80.92 ± 8.34     & 89.19 ± 5.70    & 82.09 ± 8.95      & 89.86 ± 6.12     & \textcolor{blue}{81.56} ± \textbf{6.03}        & \textcolor{blue}{89.76} ± \textbf{3.92}        \\ \hline
2                                    & 60.36 ± 7.68     & 74.99 ± 6.10    & 79.61 ± 10.23     & 88.23 ± 7.34     & \textbf{81.72 ± 6.05}        & \textbf{89.81 ± 3.93}        \\ \hline
3                                    & 59.94 ± 7.78     & 74.65 ± 6.20    & 77.57 ± 11.96     & 86.79 ± 8.62     & \textbf{81.23 ± 6.09}        &\textbf{89.53 ± 3.98}        \\ \hline
4                                    & 59.63 ± 7.94     & 74.40 ± 6.30    & 74.65 ± 12.58     & 84.83 ± 9.03     & \textbf{80.56 ± 6.04}        & \textbf{89.10 ± 3.96}       \\ \hline
5                                    & 58.51 ± 8.43     & 73.46 ± 6.83    & 72.64 ± 11.45     & 83.62 ± 8.15     & \textbf{80.59 ± 5.99}        &\textbf{89.12 ± 3.91}        \\ \hline
\end{tabular}
\end{adjustbox}

\end{center}
\label{tab:ISICAS24_alg_performance}
All values presented in the table are expressed as percentages. The evaluation results are shown as $mean $ ± $standard \ deviation$
\end{table*}

\section{Benchmark Ground Segmentation Algorithms Using SemanticKITTI Dataset}
%The advent of multi-GPU training has revolutionized the speed at which complex neural networks can be trained by distributing computational tasks across multiple devices. 
This section aims to investigate the feasibility and impact of parallel processing on ground segmentation in rotating mechanical Lidar point clouds

% Specifically, in this section, we introduce how to slice and parallel the two different types Lidar data. 

\subsection{Dataset and Metric}
The SemanticKITTI dataset \cite{behley2019semantickitti}, recognized as a benchmark in semantic scene understanding for autonomous driving applications, underpins our experimental framework. This dataset provides densely annotated, point-wise labels for Lidar point cloud sequences captured using a Velodyne HDL-64E mechanical Lidar system.  For ground segmentation, we consolidate several ground-related semantic labels—such as road, sidewalk, parking, and other ground types—into a unified ground truth category. To assess the effectiveness of the ground segmentation methods within our parallel processing framework, we employ two key evaluation metrics: Intersection over Union (IoU) and F1 Score.

\subsection{Ground Segmentation Algorithms}
To thoroughly evaluate the impact of parallelism on Lidar ground segmentation, we selected three distinctive approaches, each representing a To comprehensively evaluate the impact of parallelism on Lidar ground segmentation, we selected three distinct approaches, each representing a different algorithmic category: point-based, grid-based, and range image-based methods. Specifically, we utilized RANSAC \cite{fischlerRandomSampleConsensus1981} for point-based segmentation, the Simple Morphological Filter (SMRF) \cite{pingelImprovedSimpleMorphological2013} for grid-based segmentation, and the depth ground segmentation method (elevation angle-based) \cite{bogoslavskyi2017efficient} for range image-based segmentation. These implementations were chosen to assess the efficiency and applicability of parallel processing techniques across various segmentation strategies. Through these experiments, our study aims to illuminate the potential benefits and limitations of parallel processing when applied to different Lidar ground segmentation algorithms.

\subsection{Frame Segmentation for Mechanical Lidar}
\begin{figure}[H]
    \includegraphics[width=\linewidth]{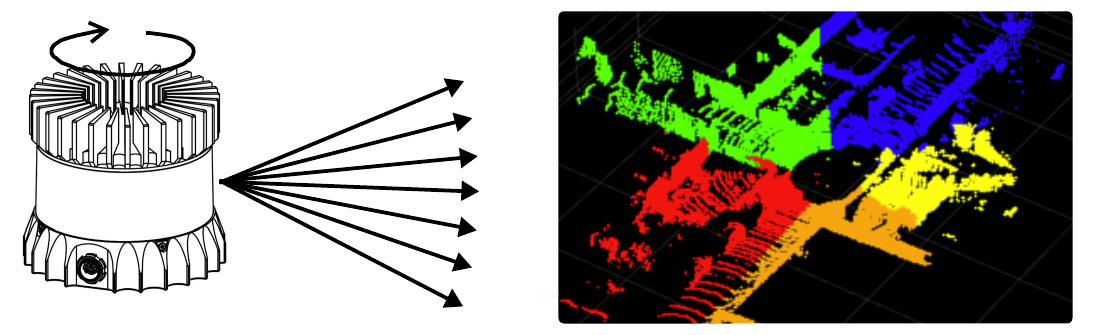}
    \caption{Mechanical Lidar Mechanism and Frame Segmentation}
    \vspace{-0.4cm}
    \label{fig:ISICAS24_2_frame_slice_mechanicall}
\end{figure}

Figure. \ref{fig:ISICAS24_2_frame_slice_mechanicall} illustrates the frame slicing methodology, showing how the beam transmitter and rotating mirror of mechanical Lidar systems scan the field of view (FoV). After applying spherical projection, we segment the Lidar data frame into horizontal angular ranges. This process involves dividing the range image into equal slices based on predetermined segments to evaluate the impact of slice size and number on ground segmentation performance. 
% In our experiments, we segmented the frame into up to five slices. These results provide valuable insights into the scalability and performance impact of parallel processing on Lidar ground segmentation.

% \textbf{Intersection over Union (IoU)}: Also known as the Jaccard Index, IoU measures the overlap between the predicted segmentation and the ground truth. For point clouds, it is defined as the ratio of the number of points correctly classified (intersection) to the total number of points across both the predicted and ground truth segments (union), excluding the intersection points counted twice.

% \textbf{F1 Score} This metric serves as the harmonic mean of precision and recall, balancing the two to provide a comprehensive measure of segmentation accuracy. Precision denotes the ratio of correctly predicted positive points to all points predicted as positive, while recall represents the ratio of correctly predicted positive points to all actual positive points. The F1 Score is calculated using the formula:$F1 = 2 \times \frac{precision \times recall}{precision + recall}$, where precision is defined as \$frac{TP}{TP + FP}$ and recall as \$frac{TP}{TP + FN}$, with TP, FP, and FN representing true positives, false positives, and false negatives, respectively.

\subsection{Result Analysis}

In our experimental framework, we divided the Lidar data frame into segments ranging from one to five slices. This segmentation aimed to explore the relationship between the number of slices and the effectiveness of different ground segmentation methods. Our goal was to identify the most efficient strategies for parallel processing, thereby enhancing computational efficiency and segmentation accuracy.

As shown in Table. I, A notable trend observed in our study is the performance degradation associated with an increasing number of slices, particularly evident in the SMRF method. As a grid-based approach, SMRF relies on constructing a minimum elevation surface map for initial point cloud segmentation, and its performance is significantly affected by variations in the area's height distribution. These variations become more pronounced when the data is segmented, especially in a bird's-eye view. Similarly, the performance of the RANSAC method deteriorates as additional slices are introduced. This decline is attributed to RANSAC's dependence on the volume of input points for accurate statistical plane fitting, which becomes less feasible as segmentation increases.

Contrary to our expectations, the range image-based Depth Ground method not only maintained robust performance across varying slice numbers but also showed improved metrics when the number of slices increased from one to two. We attribute this outcome to the fact that the vertical frame slicing has minimal impact on the label propagation process, which operates in a bottom-up manner.

Among all the methods evaluated, the Depth Ground method proved to be the most resilient and consistent, particularly excelling at handling data and tasks segmented into multiple slices. This method's robustness highlights the inherent advantages of range image-based approaches in maintaining high accuracy across diverse segmentation conditions. The point-based RANSAC method also demonstrated commendable resilience, though to a lesser degree, while the grid-based SMRF method was notably more vulnerable to the negative effects of the slicing strategy.

\section{Architecture and Implementation for Solid-State Lidar }

Building on previous experiments, the range image-based method has demonstrated superior performance and robustness, especially when combined with parallel processing for ground segmentation tasks using mechanical Lidar. In this section, we introduce the parallel range image-based ground segmentation method along with its hardware accelerator, optimized specifically for MEMS solid-state Lidar (SSL).

\subsection{Frame Segmentation for SSL}

\begin{figure}[h]
    \includegraphics[width=\linewidth]{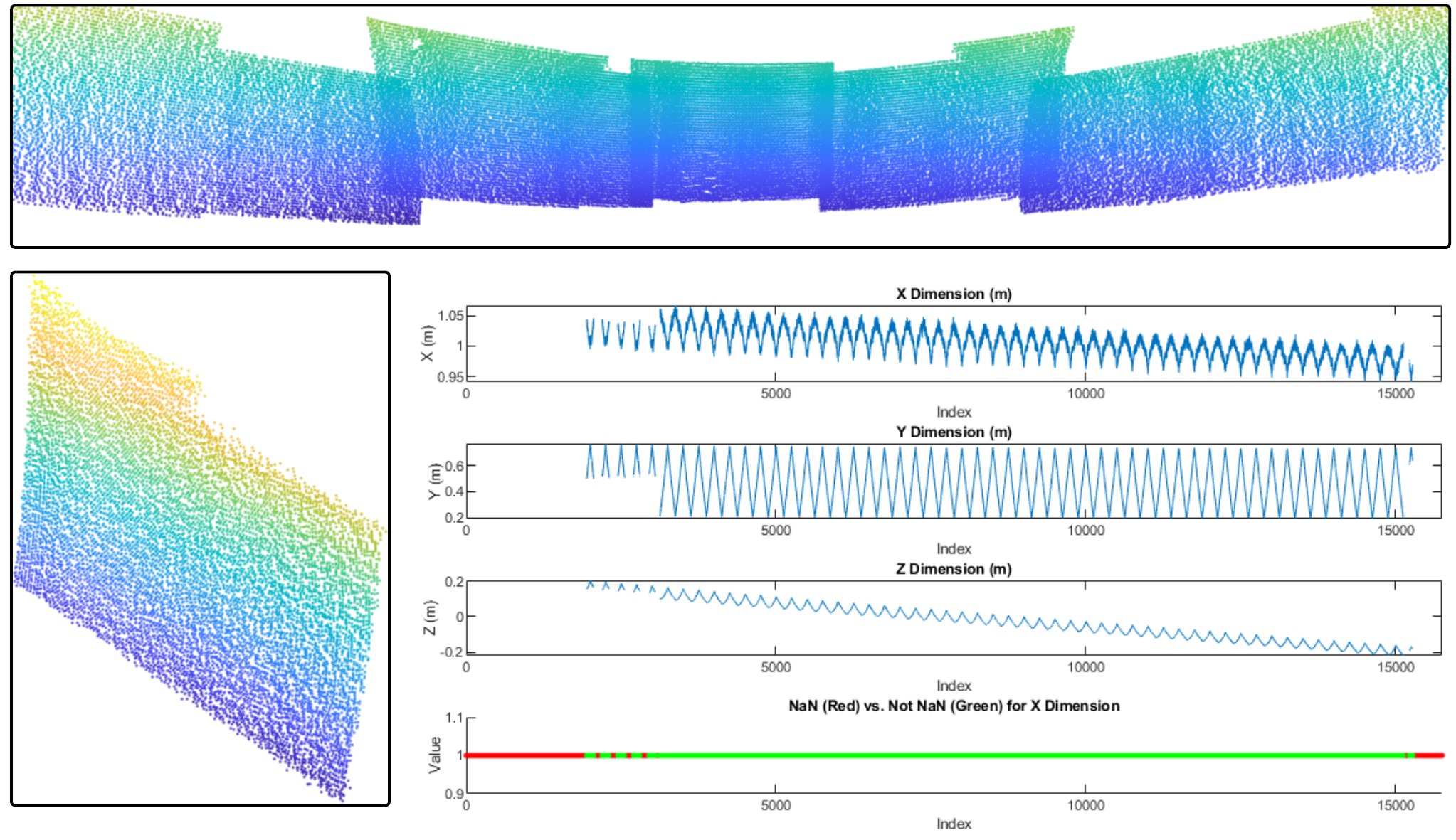}
    \caption{\textbf{Top}: The demonstrate data frame, captured using the SSL (Robosense M1) facing a flat wall, consists of a total input point sequence of 78,750 points. Upon observation, each subframe is slightly misaligned, with adjacent subframes overlapping at the edges. Additionally, some points are missing due to transmitter or receiver errors, as the Lidar sensor used was trial equipment. \textbf{ Bottom Left}: Illustration of the second subframe from the left, showing the uniform rectangular format of all the subframes. \textbf{Bottom Right}: Statistical analysis of values in the x, y, z, and validity dimensions, clearly depicting the subframe’s uniform rectangular shape with dimensions of 126 x 125.}
    \vspace{-0.4cm}
    \label{fig:ISICAS24_2_real_frame_analysis}
\end{figure}

As shown in Figure. \ref{fig:ISICAS24_2_real_frame_analysis}, the raw SSL data frame is organized in a fixed structure, consisting of 78,750 points in a single column. Upon further analysis, we determined that the data frame was divided into five uniform subframes, each containing 15,750 points, as depicted in the figure. Notably, the points within the subframes were arranged in a zig-zag pattern. To address this, we reorganized the even rows and restructured each subframe into a 126x125 matrix.

\begin{figure}[H]
    \includegraphics[width=\linewidth, height=2.5cm]{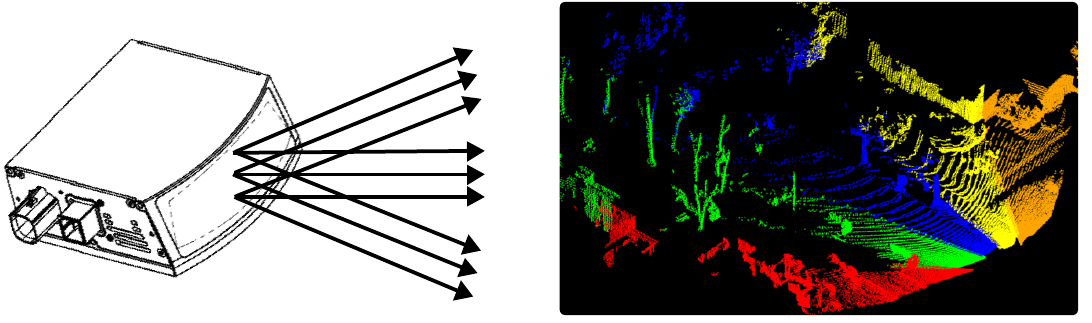}
    \caption{SSL Lidar Mechanism and Frame Segmentation}
    \vspace{-0.4cm}
    \label{fig:ISICAS24_2_frame_slice_ssl}
\end{figure}

Given the SSL data frame's natural division into five distinct parts, and to prevent the discontinuous overlap in edge areas from affecting the processing algorithms, we implemented a five-slice parallel segmentation strategy. Based on the conclusions from the previous section, each subframe underwent range image-based ground segmentation. The segmented SSL frame is illustrated in Figure. \ref{fig:ISICAS24_2_frame_slice_ssl}.

\subsection{Parallel Ground Segmentation Algorithm for SSL}
We utilize the refined depth ground segmentation (RDG) method \cite{zhangStreamBasedGroundSegmentation2024} as the backbone for implementing the parallel mechanism. RDG is a hardware-friendly approach optimized for stream-pipelined processing. The key idea is to parallelize RDG for each subframe while pipelining the propagation phase within the parallelized processing. For detailed algorithmic steps, please refer to Algorithm. \ref{alg:paralGroundSeg}.

\begin{algorithm}[H]
\caption{Parallel Ground Segmentation Algorithm}
\begin{algorithmic}[1]

\STATE \textbf{Input:} $F$, the array of organized SSL data frame
\STATE \textbf{Init:}  $L$, the array for output label masks \\ 
       \textbf{Init:}  numIter, the iteration number for label propagation

\FOR{\textbf{:: parallel} each subframe $f$ in $F$ with index $i$ }
    \STATE $f$ $\gets$ frameRepair($f$)
    \STATE $m_{\alpha}$ $\gets$ elevationMatrixCompute($f$)
    \STATE $s$ $\gets$ initSeed($m_{\alpha}$)
    \FOR{\textbf{:: pipeline} each iterator $j$ $<$ numIter}
         \STATE $l$ $\gets$ labelPropagation($s$)
    \ENDFOR
    \STATE $L$ [i] $\gets$ $l$
\ENDFOR

\STATE \textbf{Output: $L$} 
\end{algorithmic}
\label{alg:paralGroundSeg}
\end{algorithm}

\subsection{Parallel Processing Architecture}
\begin{figure}[h]
    \includegraphics[width=8.5cm, height = 2.8cm]{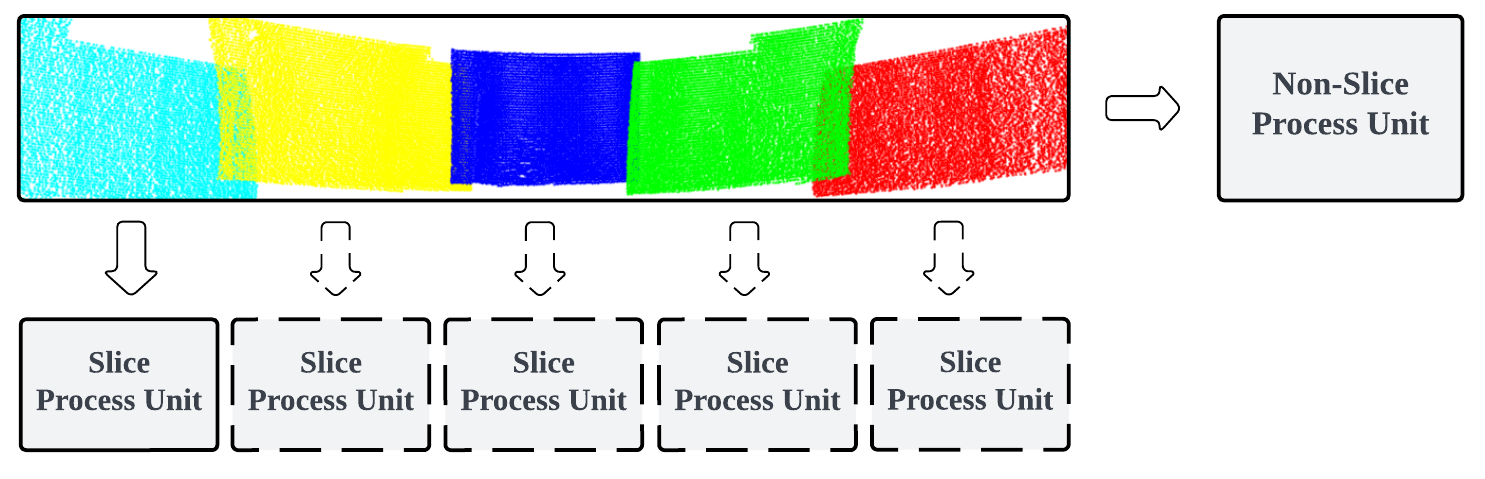}
    \caption{SSL Processing Acceleration Architectures: Implementation across different processing unit allocation strategies}
    \vspace{-0.4cm}
    \label{fig:ISICAS24_2_hardware_struct}
\end{figure}

As shown in Figure. \ref{fig:ISICAS24_2_hardware_struct},  we aim to showcase and validate the advantages of our parallel processing workflow through FPGA implementations. We conducted comparative evaluations between the non-slice configuration and the five-slice configuration, using both a single processing unit (PU) and multiple PUs.

The processing unit implementation, adapted from the refined depth ground segmentation architecture \cite{zhangStreamBasedGroundSegmentation2024}, incorporates a streamlined value smoothing module for elevation smoothing within the angular matrix and an optimized one-pass pipelined data propagation module. For further implementation details, please refer to the simplified module provided in our early contribution to the MathWorks Example.\footnote{\url{https://www.mathworks.com/help/visionhdl/ug/lidar-ground-segmentation.html}}.

\subsection{Hardware Implementation Results}

We implemented our design both with and without data frame slicing, as well as across different levels of parallel computation. This setup was intended to demonstrate the resource efficiency and processing speed benefits of parallel mechanisms on FPGA platforms. The implementation was realized using MATLAB's HDL Coder and Vision HDL Toolbox, specifically with MATLAB 2023b and Xilinx 2022.1, targeting the Xilinx xc7z045 FPGA platform.

\begin{table}[ht]
\label{tab:ISICAS24_2_arch_resource}
\begin{center}
\caption{Resource Usage of Non-Parallel, Single-Parallel, and Multi-Parallel Ground Segmentation FPGA Implementation}
 \begin{adjustbox}{width=8.5cm, center}
\begin{tabular}{c|c|c|c}
\hline
\textbf{Resource} & \textbf{Non-slice PU} & \textbf{Five-slice Single PU} & \textbf{Five-slice Multiple PUs} \\ \hline \hline
LUTs        & 32350 (14.8 \%)       & 31783 (14.54\%)   & 100060 (45.78\%)                  \\ \hline
Registers   & 36472 (8.34\%)        & 35448 (8.11\%)    & 105190 (24.06\%)                  \\ \hline
DSPs              & 26 (2.89\%)           & 26 (2.89\%)       & 78 (8.67\%)                 \\ \hline
Block RAM Tile    & 136 (24.95\%)         & 117.5 (21.56\%)   & 335 (61.47\%)               \\ \hline
\end{tabular}      
\end{adjustbox}
\end{center}
The non-slice PU processes the entire SSL data frame, which has a size of 126x625. In contrast, the five-slice single PU processes five subframes sequentially, each with a size of 126x125. The five-slice multiple PU configuration deploys three processing units to parallelize the computation. All designs successfully met the frequency constraint of 167.54 MHz.
\end{table}

The results of our comparative analysis, detailed in Table. II, reveal that the five-slice processing unit outperforms its non-slice counterpart in terms of resource utilization. This improvement is primarily attributed to the reduced row length in each subframe, which lowers the storage and processing requirements for intermediate computation.

On the other hand, increasing processing throughput through parallelization requires additional hardware resources.

\begin{figure}[h]
\begin{center}

    \includegraphics[width=8cm]{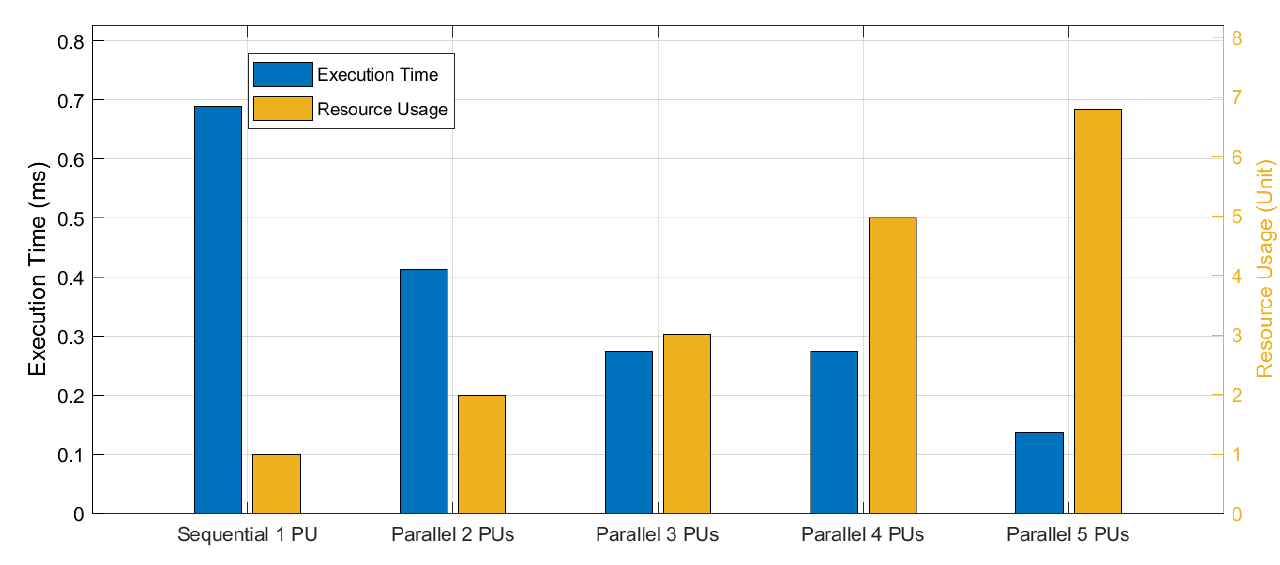}
    \caption{Resource Usage and Execution Time Across Different Levels of Parallelism: execution time is measured in ms, and resource usage value is presented as normalized values relative to a single processing unit. When the design works at 167.54 MHz, the estimated exuection time for a single PU setup is 0.137ms}
    \vspace{-0.4cm}
    \label{fig:ISICAS24_2_hardware_time_resource}
    
\end{center}
\end{figure}

As depicted in Figure. \ref{fig:ISICAS24_2_hardware_time_resource}, the relationship between resource usage and execution time is analyzed across varying levels of parallelism in the ground segmentation task. Execution time, measured in milliseconds (ms), and resource usage, normalized to a single processing unit (PU), are presented for configurations ranging from sequential processing with single PU to parallel processing with up to five PUs.

The balance point is achieved with three PUs, where execution time is significantly reduced with a reasonable increase in resource usage. While not necessarily the optimal setup, three PUs provide a recommended balance, optimizing performance without excessive resource consumption. Beyond this point, further reductions in execution time result in substantially higher resource costs, making the three PUs configuration a practical choice for balanced efficiency.

% Our findings demonstrate the inherent advantages of adopting a parallel processing approach for ground segmentation tasks, particularly when implemented on FPGA platforms. The five-slice parallel workflow exemplifies how segmentation tasks can be optimized for speed and resource usage, highlighting the potential for significant efficiency gains in real-world applications.

\section{Real-world Experiment}

\begin{figure*}[b]
    \begin{center}
    \includegraphics[width=15cm ]{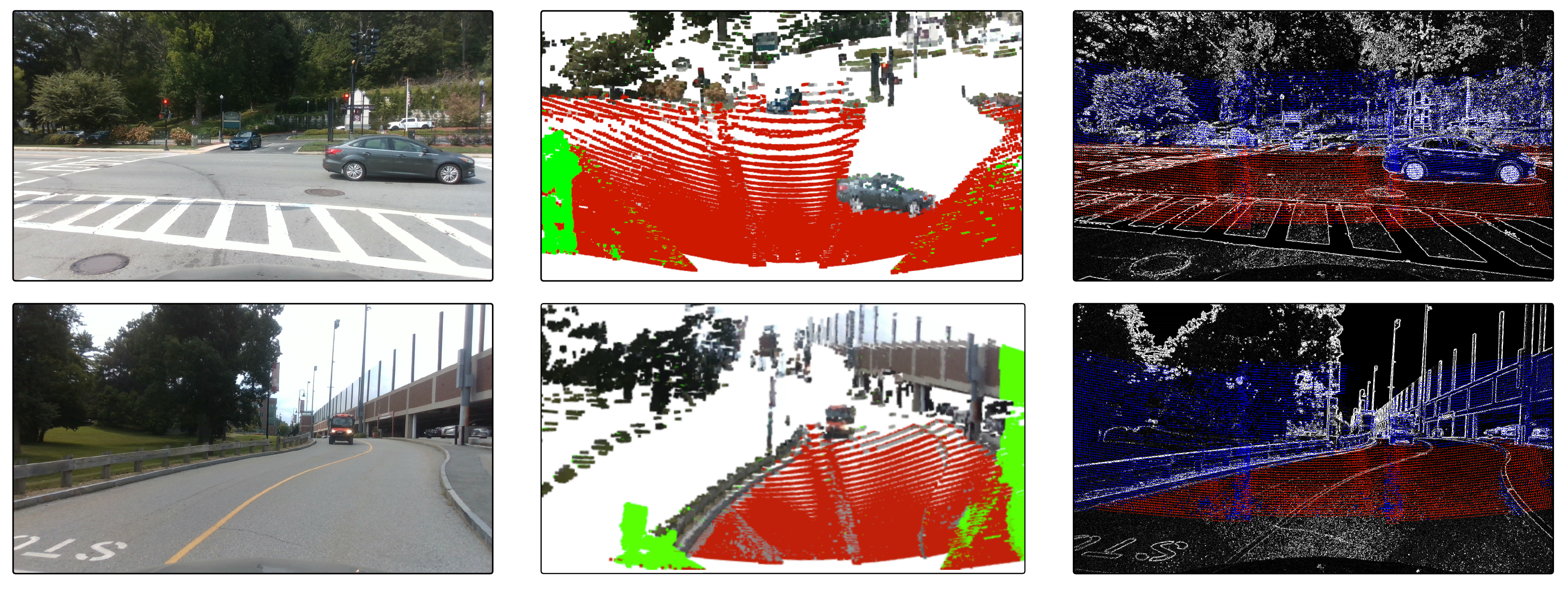}
    \caption{Segmentation Results of Plane Ground: \textbf{The left} image shows the original camera image for reference. \textbf{The middle} image displays the result in 3D space, where red points represent labeled ground and green points indicate areas outside the camera’s field of view without RGB color. \textbf{The right} image presents the result in 2D, with blue pixels denoting non-ground points and red pixels representing ground points.}
    \vspace{-0.4cm}
    \label{fig:ISICAS24_2_real_frame_quality}
     \end{center}
\end{figure*}

\begin{figure}[H]
    \includegraphics[width=\linewidth]{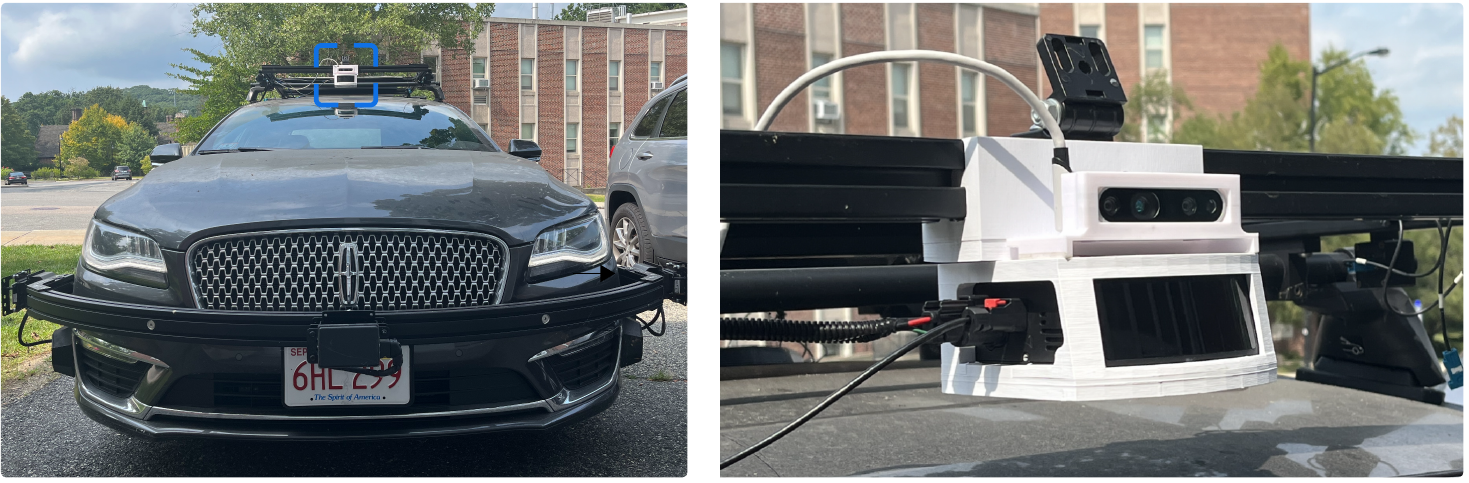}
    \caption{Data Collection Framework \textbf{Left}: The modified Lincoln MKZ vehicle equipped the camera-SSL sensing system. The system is mounted on top using a custom 3D-printed frame. \textbf{Right}: Calibrated camera-SSL sensing system combining a Robosense M1 SSL (FoV: $120^\circ\times25^\circ$) and a RealSense D435i camera (FoV: $86^\circ\times57^\circ$).}
    \vspace{-0.4cm}
    \label{fig:ISICAS24_2_real_collect}
\end{figure}

\subsection{Data Preparation}
To evaluate the efficacy and compatibility of our parallel ground segmentation algorithm-hardware co-design for solid-state Lidar sensors, we established a camera-SSL recording framework as Figure. \ref{fig:ISICAS24_2_real_collect}. Utilizing a time synchronized method with the Robot Operating System (ROS), we ensured coherent data collection between the Lidar and camera frames. Sensor calibration was performed using our previously developed method \cite{huang2024calib}, optimizing the alignment and synchronization of data streams, which enhanced the accuracy of our real-world testing.

For the performance evaluation of ground segmentation algorithms, we labeled ground points using a semi-automatic approach. Initial ground points were identified through a traditional method and then refined by referencing the corresponding camera images using MATLAB’s Lidar Labeler tool. The dataset was balanced by selecting subframes that included both flat roads and curved roads to ensure a comprehensive evaluation.

% \begin{figure}[h]
%     \includegraphics[width=\linewidth]{figures/ISICAS24_2_explore_slice_3.pdf}
%     \caption{Mechanical Lidar Slice Demonstration: For example, if slice number is five, each sub frame has a horizontal angular range of 72 degree.}
%     \vspace{-0.4cm}
%     \label{fig:ISICAS24_2_explore_slice_3}
% \end{figure}

\subsection{Quantitative Evaluation}

\begin{table}[ht]
\label{tab:ISICAS24_2_arch_resource}
\begin{center}
\caption{Performance of Ground Segmentation Methods for solid-state Lidar}
 % \begin{adjustbox}{width=5cm, center}
\begin{tabular}{c|c|c|c}
\hline
\textbf{Method}      & \textbf{IoU} & \textbf{F1}    & \textbf{Run Time} \\ \hline \hline
SMRF                 & 61.44        & 75.25          & 482.38             \\ \hline
RANSAC               & 74.85       & 84.92          & 12.07              \\ \hline
Depth Ground         & 91.71        & 95.57          & 4.31              \\ \hline
Parallel Ground on FPGA* &    \textcolor{blue}{89.10}    & \textcolor{blue}{94.14} & \textbf{0.28}       \\ \hline
\end{tabular}   
% \end{adjustbox}
\end{center}
All performance values presented in the table are expressed as percentages, the time values are expressed as milisecond (ms) *The FPGA implementation includes three PUs @ 167.54MHz. Other methods are running through the MATLAB implementation on Intel i7-12700K CPU @ 3.6GHz
\end{table}

Table. III compares the IoU, F1 scores, and runtime for various ground segmentation methods applied to solid-state Lidar data, including SMRF, RANSAC, Depth Ground, and our proposed parallel ground method on FPGA.

The Depth Ground method on CPU and the parallel ground method on FPGA stand out as the most effective approaches, achieving high IoU and F1 scores. Moreover, the parallel ground method on FPGA offers a significant advantage in runtime efficiency, completing the segmentation task in just 0.28 ms—a substantial improvement compared to the other methods. Even the relatively fast Depth Ground method requires 4.3 seconds. Our proposed method is approximately 15.4 times faster. Additionally, the power consumption for the three-PU setup is only 3.52 W, as reported in the implementation.

Notably, if the parallel processing units (PUs) were extended to five, the maximum potential speedup could reach approximately 30.9 times than the depth ground method, further enhancing the efficiency of the method. This scalability demonstrates the method's capability to handle even more demanding real-time applications.

Overall, the parallel ground method on FPGA is not only among the top two in terms of accuracy but is also the most efficient, offering significant speedup potential with additional parallel units. This makes it highly suitable for real-time edge Lidar processing tasks.

\subsection{Qualitative Evaluation}

Figure. \ref{fig:ISICAS24_2_real_frame_quality} showcases the segmentation results of plane ground detection across different representations: the original camera image (left), the segmentation result in 3D point cloud space (middle), and the result in 2D image space (right). The original camera image provides a visual reference of the scene. The middle image, showing the colored 3D point cloud with accurate calibration. The colored points can clearly indicating the space position of labeled data. In the right image, the segmentation results in 2D image space show that the labeled ground plane is clearly distinguished from other elements.

In the top image pairs, the ground is primarily a flat road, showing accurate segmentation. In the bottom pairs, the ground includes some curvatures near the fence. The detected plane’s edge aligns closely with the actual edge, demonstrating the method’s sensitivity to plane curvature. However, there is some mislabeling in the right bottom corner, where the subframe is entirely above the curb. The overlap between subframes also contributes to minor segmentation errors.

Additionally, the range-image-based method demonstrates higher accuracy compared to the SemanticKITTI data using mechanical Lidar. This improvement is attributed to the denser and more evenly distributed SSL data, which benefits from the direct reception of the scan frame by MEMS laser arrays which results in better frame quality with fewer missing points between vertical channels. Differences in ground labeling and recording environments may also contribute to these results. The RANSAC method performed lower, as we balanced the evaluation between flat ground and curved surfaces. Meanwhile, SMRF tends to segment all lower points, but not necessarily just the ground, leading to a lower evaluation score.

% \begin{figure*}[b]
%     \begin{center}
%     \includegraphics[width=15cm ]{figures/ISICAS24_2_real_frame_quality.pdf}
%     \caption{Segmentation Result of Plane Ground: Left, result in image space. Right: result in point cloud space}
%     \vspace{-0.4cm}
%     \label{fig:ISICAS24_2_real_output_small}
%      \end{center}
% \end{figure*}

\section{Conclusion} 
In this research, we have successfully demonstrated the advantages of parallel processing for Lidar ground segmentation through a comprehensive comparison of existing algorithms, the design of a scalable acceleration architecture, and real-world validation. Our analysis using the SemanticKITTI dataset highlighted the superior performance of the range-image-based segmentation method, which remained robust even under frame-slicing, outperforming other methods in terms of both accuracy and stability. Furthermore, we developed a scalable FPGA-based ground segmentation accelerator tailored for solid-state Lidar, which exhibited resource efficiency, high throughput, and low power consumption, making it adaptable to various system constraints. Finally, the implementation and validation of our approach using a custom camera-SSL dataset on a test vehicle confirmed its real-world effectiveness, achieving significant speedups—up CPU methods, while maintaining segmentation accuracy. These findings validate the feasibility of parallel processing for point cloud ground segmentation and highlight a promising direction for future solid-state Lidar systems.

%%%%%%%%%%%%%%%%%%%%%%%%%%%%%%%%%%%%%%%%%%%%%%%%%%%%%%%%%%%%%%%%%%%%%%%%%%%%%%%%

%%%%%%%%%%%%%%%%%%%%%%%%%%%%%%%%%%%%%%%%%%%%%%%%%%%%%%%%%%%%%%%%%%%%%%%%%%%%%%%

%%%%%%%%%%%%%%%%%%%%%%%%%%%%%%%%%%%%%%%%%%%%%%%%%%%%%%%%%%%%%%%%%%%%%%%%%%%%%%%%

\bibliographystyle{IEEEtran}
\bibliography{IEEEabrv,IEEEexample}

\end{document}